\documentclass[letterpaper, 10 pt, conference]{ieeeconf}  

\IEEEoverridecommandlockouts                              

\usepackage{graphics} 
\usepackage{epsfig} 
\usepackage{mathptmx} 
\usepackage{times} 
\usepackage{amsmath} 
\usepackage{amssymb}  

\usepackage{algorithm}
\usepackage{algorithmic}

\usepackage{amsfonts}
\usepackage{amsmath}

\usepackage{booktabs}

\usepackage{multirow}

\usepackage{bbding}

\usepackage{makecell}

\usepackage{adjustbox}
\newcolumntype{R}[2]{%
    >{\adjustbox{angle=#1,lap=1.3\width-(#2)}\bgroup}%
    l%
    <{\egroup}%
}

\usepackage{cite}

\usepackage{hyperref}

\title{\bf Domain Adaptation-Based \\
Crossmodal Knowledge Distillation for 3D Semantic Segmentation}

\author{
    \authorblockN{Jialiang Kang, Jiawen Wang, and Dingsheng Luo, \IEEEmembership{Member, IEEE}}
    \authorblockA{School of Intelligence Science and Technology, Peking University}
}

\begin{document}

\maketitle
\thispagestyle{empty}
\pagestyle{empty}




\begin{abstract}
        Semantic segmentation of 3D LiDAR data plays a pivotal role in autonomous driving. Traditional approaches rely on extensive annotated data for point cloud analysis, incurring high costs and time investments. In contrast, real-world image datasets offer abundant availability and substantial scale. To mitigate the burden of annotating 3D LiDAR point clouds, we propose two crossmodal knowledge distillation methods: \textbf{U}n\-su\-per\-vised \textbf{D}o\-main \textbf{A}d\-ap\-ta\-tion \textbf{K}nowl\-edge \textbf{D}is\-til\-la\-tion (UDAKD) and \textbf{F}ea\-ture and \textbf{S}e\-man\-tic-based \textbf{K}nowl\-edge \textbf{D}is\-til\-la\-tion (FSKD). Leveraging readily available spatio-temporally synchronized data from cameras and LiDARs in autonomous driving scenarios, we directly apply a pretrained 2D image model to unlabeled 2D data. Through crossmodal knowledge distillation with known 2D-3D correspondence, we actively align the output of the 3D network with the corresponding points of the 2D network, thereby obviating the necessity for 3D annotations. Our focus is on preserving modality-general information while filtering out modality-specific details during crossmodal distillation. To achieve this, we deploy self-calibrated convolution on 3D point clouds as the foundation of our domain adaptation module. Rigorous experimentation validates the effectiveness of our proposed methods, consistently surpassing the performance of state-of-the-art approaches in the field. Code is available at \texttt{\url{https://github.com/KangJialiang/DAKD}}.
\end{abstract}

\section{Introduction}
\label{sec:Introduction}

Recent years have witnessed a surge in perception algorithms for processing LiDAR point cloud data, supported by diverse open benchmarks~\cite{9,10,11,12}. Despite these advances, the efficacy of such techniques is often impeded by their reliance on meticulously annotated data, leading to time-consuming and labor-intensive annotation procedures~\cite{41,51}. Moreover, existing domain adaptation methods have shown limited effectiveness in handling the complexities of 3D point clouds, necessitating separate training of 3D networks for different LiDAR configurations, thereby exacerbating the annotation burden~\cite{102}. Consequently, a pressing demand has arisen to address the challenge of annotating LiDAR point clouds.

\begin{figure*}[t]
        \centering
        \includegraphics[scale=1]{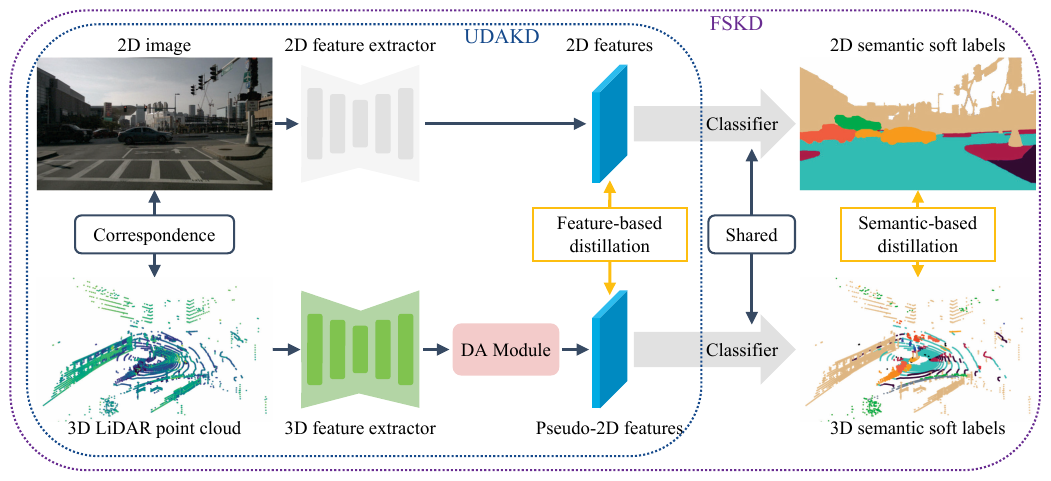}
        \caption{Illustration of UDAKD and FSKD. Features are generated from 3D point clouds and 2D images through their respective networks. The 3D features are input into the domain adaptation (DA) module and aligned with the 2D feature extractor outputs. In FSKD, both 2D and 3D features are additionally fed into the same multilayer perceptron (MLP)-based classifier to obtain point-wise semantic soft labels and conduct semantic-based knowledge distillation.}
        \label{fig:overview}
\end{figure*}

To reduce the annotation burden while enhancing semantic segmentation learning for 3D networks, we employ crossmodal knowledge distillation from 2D to 3D, leveraging paired image and LiDAR data. We introduce two methods: \textbf{U}n\-su\-per\-vised \textbf{D}o\-main \textbf{A}d\-ap\-ta\-tion \textbf{K}nowl\-edge \textbf{D}is\-til\-la\-tion (UDAKD) and \textbf{F}ea\-ture and \textbf{S}e\-man\-tic-based \textbf{K}nowl\-edge \textbf{D}is\-til\-la\-tion (FSKD), as depicted in Fig.~\ref{fig:overview}.

In UDAKD, akin to PPKT~\cite{48} and SLidR~\cite{51}, we bypass the need for the original 2D training dataset by directly aligning 2D and 3D network features on multimodal data, providing a pretraining approach for the 3D network.

In FSKD, we leverage the 2D semantic segmentation model trained on labeled images to extract features and generate per-pixel soft labels for unlabeled target data. By aligning the features and labels from the 3D network with the corresponding 2D network points, we train a comprehensive 3D semantic segmentation model without explicit 3D annotations.

Inspired by~\cite{54}, we aim to minimize modality-specific feature interference while maximizing modality-general feature extraction during crossmodal knowledge distillation. To achieve this, we design a domain adaptation module based on self-calibrated convolution~\cite{62}, which has shown promise in aligning feature distributions of images with those from LiDAR point clouds~\cite{53}. We implement 3D self-calibrated convolution within the LiDAR point cloud modality.

Our main contributions can be summarized as follows:

\begin{itemize}
        \item We propose a novel domain adaptation module that employs 3D self-calibrated convolution to transfer modality-general features and eliminate modality-specific details.

        \item We introduce distinct crossmodal knowledge distillation strategies for unlabeled and labeled images, enhancing 3D semantic segmentation via image-to-LiDAR knowledge distillation.

        \item Our UDAKD and FSKD methods, designed for unlabeled and labeled image scenarios respectively, achieve superior performance compared to previous state-of-the-art (SOTA) methods.
\end{itemize}

\section{Related Work}
\label{sec:Related Work}

\subsection{LiDAR-Based Semantic Segmentation}
\label{LiDAR-Based Semantic Segmentation}

Semantic segmentation networks for LiDAR data predominantly rely on the U-Net architecture~\cite{17}. Given the sparsity of point clouds, three main representations are used: \textbf{i) Point-based} methods process original points in continuous 3D space~\cite{18,19,21,22}, often suitable for small-scale synthetic or indoor point clouds. \textbf{ii) Projection-based} methods project point clouds onto range views~\cite{27,28,29,30,31} or bird's-eye views~\cite{32,33}, potentially disrupting the 3D spatial structure. \textbf{iii) Voxel-based} methods, e.g., Minkowski U-Net~\cite{35,36}, Cylinder3D~\cite{37}, AF2-S3Net~\cite{38}, SPVNAS~\cite{39}, and SVQNet~\cite{100}, quantize 3D points into sparse voxels, convolving only on non-empty voxels to reduce computational costs. Our work adopts the voxel-based approach for efficient LiDAR point cloud feature extraction.

\subsection{Crossmodal Supervision between Image and Point Cloud}

We utilize knowledge distillation for crossmodal supervision, transferring knowledge from a 2D image teacher network to a 3D point cloud student network, enhancing the 3D network's performance with limited annotations.

\paragraph{Indoor scenario} Gupta et al.~\cite{45} transfer knowledge from annotated color images to unannotated depth or optical flow images using intermediate representations. Meyer et al.~\cite{47} introduce multimodal contrastive learning for RGB-D images when only one modality is available during inference. PPKT~\cite{48} pretrains a 3D network on unlabeled RGB-D datasets using an existing 2D network without accessing the original 2D training data. I2P-MAE~\cite{I2P-MAE} uses self-supervised 2D knowledge to guide the 3D masked autoencoding process.

\paragraph{Autonomous driving scenario} Matching 2D and 3D points in autonomous driving is challenging due to sparse LiDAR sampling and continuous motion~\cite{99}. To address this, 2D3DNet~\cite{41} uses high-confidence 2D outputs as pseudo-labels and integrates weighted voting of these outputs as semantic features for the 3D network. xMUDA and xMoSSDA~\cite{42,49} enforce consistency between point cloud and image predictions through mutual mimicking. OpenScene~\cite{OpenScene} uses a 2D model pretrained for open-vocabulary semantic segmentation to supervise the 3D network. CMKD~\cite{53} transforms image features to resemble point clouds using a self-calibrated convolution-based domain adaptation module. SLidR~\cite{51} employs contrastive learning on SLIC superpixels~\cite{52} to address coverage issues in LiDAR point clouds. While previous works primarily enhance 2D data processing for improved teacher models, optimizing the 3D student has been somewhat overlooked. Our work fills this gap by integrating a domain adaptation module into the 3D student network.

\paragraph{Theoretical studies} Image2Point~\cite{55} directly transfers convolutional kernels from 2D to 3D, achieving competitive point cloud classification performance. They establish that the extent of neural collapse~\cite{56} in a model pretrained on source domain data sets the upper limit of its performance on target tasks. Inspired by~\cite{54}, which proposes the modality focusing hypothesis (MFH), we emphasize modality-general features in crossmodal knowledge distillation. Our domain adaptation module transfers modality-general features while separating modality-specific features, diverging from previous methods with fewer such considerations.

\section{Method}

\subsection{Problem Setup}

Let $\mathbf{S} = \{\mathbf{I}_s\}$ be an unlabeled dataset of RGB images from a singlemodal source domain $s$, used for self-supervised pretraining of a feature extractor $e_{\theta_e}$ with parameters $\theta_e$. When labels are available, $\mathbf{S} = \{\mathbf{I}_s, \mathbf{L}_s\}$ includes corresponding semantic annotations $\mathbf{L}_s$. In this scenario, we can pretrain a 2D semantic segmentation network $f_\theta = d_{\theta_d} \circ e_{\theta_e}$, where $\circ$ denotes function composition and $d_{\theta_d}$ is a per-pixel classifier producing soft labels for $C_{\text{cls}}$ semantic classes.

In the target domain $t$, $\mathbf{T} = \{\mathbf{I}_t, \mathbf{P}_t\}$ contains synchronized images $\mathbf{I}_t$ and point clouds $\mathbf{P}_t$. Our goal is to use $\mathbf{T}$ and the 2D extractor $e_{\theta_e}$ to obtain a 3D feature extractor $h_{\omega_h}$, or with the aid of $\mathbf{T}$ and the complete 2D network $f_\theta$, to derive a 3D segmentation network $g_\omega = k_{\omega_k} \circ h_{\omega_h}$. If annotations $\mathbf{L}_t$ are available, we refine the 3D network to $\widetilde{g}_{\widetilde{\omega}}$ with modified structure $\widetilde{g}$ and parameters $\widetilde{\omega}$.

\subsection{Framework Overview}

We propose two distinct approaches: \textbf{U}n\-su\-per\-vised \textbf{D}o\-main \textbf{A}d\-ap\-ta\-tion \textbf{K}nowl\-edge \textbf{D}is\-til\-la\-tion (UDAKD) for scenarios with unlabeled images, and \textbf{F}ea\-ture and \textbf{S}e\-man\-tic-based \textbf{K}nowl\-edge \textbf{D}is\-til\-la\-tion (FSKD) for labeled images. These methods are illustrated in Fig.~\ref{fig:overview}.

Both UDAKD and FSKD use dedicated networks to extract features from 3D point clouds and 2D images. The 3D features are adapted within the domain adaptation module $m_\lambda$ to resemble 2D features, facilitating feature-based knowledge distillation. In FSKD, 2D and 3D features are further fed into a shared MLP classifier to generate point-wise semantic soft labels, enabling semantic-based knowledge distillation.

Similar to TPVFormer~\cite{86}, our approach imposes no constraints on the number of cameras or LiDAR sensors. Distillation is performed in regions visible to both modalities.

\subsection{Segmentation Networks}

\paragraph{2D feature extractor} The 2D feature extraction network $e_{\theta_e}$ converts an RGB image $\mathcal{I}\in\mathbb{R}^{3\times H \times W}$ into $C_{\text{img}}$-dimensional image features $\mathcal{F}\in\mathbb{R}^{C_{\text{img}}\times H \times W}$:
\begin{equation}
        \label{eq:F}
        \mathcal{F} = e_{\theta_e}\left( \mathcal{I} \right).
\end{equation}
In UDAKD, $e_{\theta_e}$ comprises a ResNet-50 backbone with parameters $\theta^{\text{\tiny bck}}$ and a projection head with $1 \times 1$ convolutions and bilinear upsampling, parameterized by $\theta^{\text{\tiny hd}}$, to align with the input image size. For FSKD, we employ SegNet~\cite{68} as $e_{\theta_e}$ to reduce parameter count and computational cost while maintaining performance. Our algorithm allows flexible selection from various mature networks.

\paragraph{3D feature extractor} For 3D feature extraction, we use MinkUNet32~\cite{51,36} as $h_{\omega_h}$, which maps a point cloud $\mathcal{P}\in\mathbb{R}^{D_{}\times N_{}}$ with $N_{}$ points and $D_{}$ dimensions to features $\mathcal{G}\in\mathbb{R}^{C_{\text{pc}} \times N_{}}$ with $C_{\text{pc}}$ dimensions:
\begin{equation}
        \label{eq:G}
        \mathcal{G} = h_{\omega_h}\left( \mathcal{P} \right).
\end{equation}

\paragraph{Shared classifier} In FSKD, we use a 2D classifier $d_{\theta_d}$ and a 3D classifier $k_{\omega_k}$ alongside their respective feature extractors to form complete 2D network $f_\theta = d_{\theta_d} \circ e_{\theta_e}$ and 3D network $g_\omega = k_{\omega_k} \circ h_{\omega_h}$. These classifiers produce soft semantic labels $\mathcal{S} \in \mathbb{R}^{C_{\text{cls}} \times H \times W}$ and $\mathcal{T} \in \mathbb{R}^{C_{\text{cls}} \times N_{}}$ from features $\mathcal{F}$ and $\mathcal{G}$, respectively:
\begin{gather}
        \label{eq:S and T}
        \mathcal{S} = d_{\theta_d}\left( \mathcal{F} \right), \qquad
        \mathcal{T} = k_{\omega_k}\left( \mathcal{G} \right).
\end{gather}
For zero-shot domain adaptation, allowing the 3D network to learn new classes from images alone, we adopt a unified structure with shared parameters between $d_{\theta_d}$ and $k_{\omega_k}$, i.e., $k_{\omega_k} = d_{\theta_d}$. By training a modified 2D segmentation network $f_{\theta^\prime}^\prime = d_{\theta_d^\prime}^\prime \circ e_{\theta_e}$ on new images, keeping $e_{\theta_e}$ fixed and introducing a new classifier $d_{\theta_d^\prime}^\prime$, we derive a new 3D semantic segmentation network $g_{\omega^\prime}^\prime = d_{\theta_d^\prime}^\prime \circ m_\lambda \circ h_{\omega_h}$, capable of recognizing new classes without annotated 3D data.

Under the \emph{neural collapse} phenomenon~\cite{56}, where over-parameterized classifiers yield features clustering around class means, we assume complete neural collapse for both 2D and 3D extractors. Ideally, this allows a linear transformation to map 3D point cloud features to 2D features, with the classifier converting these into semantic labels. Using a shared MLP-based classifier for both networks thus minimizes feature extractor constraints and enables zero-shot learning for the 3D network. Acknowledging the impracticality of perfect neural collapse, we introduce a domain adaptation (DA) module to handle this variability, as detailed below.

\subsection{Domain Adaptation Module}\label{sec:Domain Adaptation Module}

Previous work on multimodal fusion typically involves both 3D point clouds and 2D images during training and inference, leading to high computational costs and slower inference due to separate network structures for each modality. Common solutions include early MLP fusion~\cite{42,43,88} or using image network outputs as additional features for 3D points~\cite{41,89,90,91}, followed by end-to-end training.

In our task, the absence of the image modality during inference requires the 3D network to independently perform semantic segmentation on point clouds without relying on image inputs. While during knowledge distillation, the 3D network learns solely from its 2D teacher. Hence, a dedicated domain adaptation module is essential.

We employ the domain adaptation module $m_\lambda$ based on self-calibrated convolution, initially proposed by~\cite{62}. Although methods like CMKD~\cite{53} and MSNetSC~\cite{92} have applied this in 2D contexts, its applicability in 3D remains unexplored. We utilize multi-layered 3D self-calibrated convolutions in $m_\lambda$ for both UDAKD and FSKD, converting 3D features $\mathcal{G}\in\mathbb{R}^{C_{\text{pc}} \times N_{}}$ into pseudo-2D features $\mathcal{G}^\prime\in\mathbb{R}^{C_{\text{img}}\times N_{}}$:
\begin{equation}
        \mathcal{G}^\prime = m_\lambda\left(\mathcal{G}\right).
\end{equation}

Previous studies~\cite{41,51} on 2D-to-3D supervision often neglect the 3D network's effort in extracting meaningful insights from limited knowledge. The DA module is crucial and may outweigh the importance of specific structural details, as it takes the job of efficiently transforming crossmodal information, allowing the 3D network to focus on learning semantically rich features. While the DA module can vary, including MLP and vanilla convolutions, self-calibrated convolutions offer superior efficiency.

\subsection{Feature-Based Knowledge Distillation}

Both UDAKD and FSKD entail aligning the distribution of 3D features $\mathcal{G}^{C_{\text{pc}} \times N_{}}$ with that of 2D features $\mathcal{F}^{C_{\text{img}}\times H \times W}$ through feature-based knowledge distillation. This process enables the 3D feature extractor $h_{\omega_h}$ to learn to transform inputs into semantically rich representations.

In the UDAKD framework, the 3D network is trained exclusively with feature-based distillation, formalized by the loss $\mathcal{L}_{\text{\tiny kd}}^{\text{\tiny UDAKD}}=\mathcal{L}_{\text{\tiny feat}}^{\text{\tiny UDAKD}}$. The main goal is to enhance knowledge transfer from the 2D teacher to the 3D student at the feature level. Our approach resembles SLidR~\cite{51}, which aligns features at the superpixel level using contrastive learning, but extends it by incorporating the domain adaptation module $m_\lambda$, leading to significant improvements. Before alignment, we average pool the 2D feature map $\mathcal{F}$ over pixels within SLIC~\cite{52} superpixels and the pseudo-2D feature map $\mathcal{G}^\prime$ over points within corresponding superpoints, yielding pooled feature vectors $\mathbf{f}_i$ and $\mathbf{g}_i$:
\begin{gather}
        \mathbf{f}_i^{} = \frac{1}{\left|\mathcal{A}_i^{}\right|} \sum_{\text{pix} \in \mathcal{A}_i^{}} \mathcal{F}(\text{pix}), \qquad
        \mathbf{g}_i^{} = \frac{1}{\left|\mathcal{B}_i^{}\right|} \sum_{\text{pnt} \in \mathcal{B}_i^{}} \mathcal{G}^\prime(\text{pnt}),
\end{gather}
where $\mathcal{A}_i^{}$ is the set of pixels in the $i$-th superpixel, $\mathcal{B}_i^{}$ is the corresponding 3D points, and $\left|\cdot\right|$ denotes the set size. Computation occurs only if $\left|\mathcal{B}_i^{}\right| > 0$. The feature distillation loss is calculated via InfoNCE~\cite{87}:
\begin{equation}
        \label{eq:l_feat_udakd}
        \begin{split}
                  & \mathcal{L}_{\text{\tiny feat}}^{\text{\tiny UDAKD}}({\theta^{\text{\tiny hd}}},{\omega_h},{\lambda})                                                                                                                                                                                      \\
                = & -\sum_{i} \log{\left[\frac{\exp{\left(\frac{{\mathbf{g}_i^{}}^\top \mathbf{f}_i^{}}{\tau}\right)}} {\sum_{i^\prime \neq i}\exp{\left(\frac{{\mathbf{g}_i^{}}^\top \mathbf{f}_{i^\prime}^{}}{\tau}\right)}+\exp{\left(\frac{{\mathbf{g}_i^{}}^\top \mathbf{f}_i^{}}{\tau}\right)}}\right]},
        \end{split}
\end{equation}
where $\tau$ is the temperature. During distillation, the parameters of the image feature extractor backbone ${\theta^{\text{\tiny bck}}}$ are frozen, while other parameters, including ${\theta^{\text{\tiny hd}}}$, ${\omega_h}$, and ${\lambda}$, are updated.

In FSKD, the feature distillation loss is the mean square error between $\mathcal{G}^\prime$ and $\mathcal{F}$:
\begin{equation}
        \mathcal{L}_{\text{\tiny feat}}^{\text{\tiny FSKD}}({\omega_h},{\lambda}) = {MSE}\left(\mathcal{G}^\prime, \mathcal{F}\right).
\end{equation}
The correspondence between coordinates in $\mathcal{G}^\prime$ and $\mathcal{F}$ is determined by extrinsic parameter calibration.

\subsection{Semantic-Based Knowledge Distillation}

In FSKD, leveraging the full 2D semantic segmentation network $f_\theta$, we conduct semantic distillation by aligning the  3D network $g_\omega$ generated soft labels $\mathcal{T}^{C_{\text{cls}} \times N_{}}$ with the output $\mathcal{S}^{C_{\text{cls}} \times H \times W}$ of $f_\theta$. The semantic distillation loss is computed using KL divergence:
\begin{equation}
        \mathcal{L}_{\text{\tiny sem}}^{\text{\tiny FSKD}}({\omega},{\lambda}) = {KL}\left(\mathcal{T} \| \mathcal{S}\right).
\end{equation}

The overall knowledge distillation loss in FSKD, $\mathcal{L}_{\text{\tiny kd}}$, is a weighted sum of feature and semantic distillation losses: $\mathcal{L}_{\text{\tiny kd}}^{\text{\tiny FSKD}} = a \mathcal{L}_{\text{\tiny feat}}^{\text{\tiny FSKD}} + b \mathcal{L}_{\text{\tiny sem}}^{\text{\tiny FSKD}}$, where $a$ and $b$ are scaling factors. Due to the combined feature-based and semantic-based distillation, superpixels and the image projection head are unnecessary for feature alignment in FSKD.

\section{Experiment}

\subsection{Datasets}

Before FSKD, a complete 2D segmentation network $f_\theta$ is trained on the singlemodal nuImages dataset, resembling scenes in nuScenes but without overlapping data. nuImages annotations cover 26 classes, mapped to a commonly used set of 11 classes, a subset of nuScenes. nuScenes serves as the main target modality dataset $\mathbf{T}$. Following~\cite{51}, 100 scenes are set aside for hyperparameter tuning, while all keyframes from the remaining 600 training scenes are used for crossmodal knowledge distillation. The original 32 classes in nuScenes are reduced to 16 by merging similar categories~\cite{51}. To validate generalization, we also fine-tune the 3D network $h_{\omega_h}$ on SemanticKITTI~\cite{44}.

\subsection{Few-Shot Semantic Segmentation}
Few-shot semantic segmentation assesses various methods for pretraining the 3D feature extractor $h_{\omega_h}$. Since each of our methods supports this task, we include both in the evaluation. We compare our methods against three strong baselines, including the top-performing predecessor SLidR~\cite{51}.

\paragraph{Training settings}\label{para:Training Settings}
In UDAKD, we directly use the ResNet-50 pretrained on ImageNet~\cite{58} with MoCo v2~\cite{64} as the 2D teacher network's backbone without fine-tuning. During knowledge distillation, we use the SGD optimizer with a momentum of 0.9, damping of 0.1, weight decay of 0.0001, and an initial learning rate of 0.5 to optimize the image projection head, 3D feature extractor, and domain adaptation module. Training lasts up to 50 epochs with a batch size of 4, employing the cosine annealing scheduler to decay the learning rate to 0 at the end of training. The temperature $\tau$ in (\ref{eq:l_feat_udakd}) is set to 0.07. Each image is partitioned into a maximum of 150 SLIC superpixels.

Before conducting FSKD, $f_\theta$ is trained on the nuImages dataset for 100 epochs with cross-entropy and Lovász loss. During the distillation phase, we freeze $f_\theta$ and optimize the 3D feature extractor $h_{\omega_h}$ and the domain adaptation module $m_{\lambda}$ using SGD. The momentum is set to 0.9, damping to 0.1, weight decay to 0.0001, and the initial learning rate to 0.05 for $h_{\omega_h}$ and 5e-5 for $m_{\lambda}$, with a batch size of 16. Training is conducted for a maximum of 50 epochs using the cosine annealing scheduler. The ratio of $\mathcal{L}_{\text{\tiny feat}}$ to $\mathcal{L}_{\text{\tiny sem}}$ is set at 10:1.

\paragraph{Baselines}
We compare our methods with I2P-MAE~\cite{I2P-MAE}, PPKT~\cite{48}, and SLidR~\cite{51}. I2P-MAE is a sophisticated masked autoencoding framework for image-to-point knowledge transfer. We include it to evaluate the adaptability of methods excelling in dense synthetic or indoor point cloud contexts to the challenges of sparse point cloud scenarios in autonomous driving. Our implementations of PPKT and SLidR are based on the publicly available code from~\cite{51}.

\paragraph{Evaluation results}
We evaluate the mIoU achieved by various methods on nuScenes and SemanticKITTI validation sets after fine-tuning on $1\%$ or $100\%$ annotated scenes from their respective training sets. In the fine-tuning phase, the 3D feature extractor parameters $\omega_h$, obtained through diverse methods, initialize $h_{\widetilde{\omega_h}}$. Simultaneously, a classifier $\widetilde{k}_{\widetilde{\omega_k}}$, trained from scratch, amalgamates to form the 3D segmentation network $\widetilde{g}_{\widetilde{\omega}} = \widetilde{k}_{\widetilde{\omega_k}} \circ h_{\widetilde{\omega_h}}$. Results are presented in Table~\ref{tab:few-shot}, where the Random baseline denotes random initialization of $h_{\widetilde{\omega_h}}$. Pretraining with I2P-MAE underperforms due to masking much of the already sparse point cloud. Our UDAKD approach extends SLidR by adding a domain adaptation module $m_{\lambda}$ with a 3-layer self-calibration convolution. Including $m_{\lambda}$ improves mIoU by 1 percentage point, demonstrating its effectiveness.

In FSKD, we utilize a 2D network trained on labeled image data, unlike other methods that use a frozen ResNet-50 pretrained via self-supervision. Although this approach might seem biased, the effectiveness of self-supervision relies on balanced, high-quality datasets~\cite{58,103}, often curated by human annotators. By training the image network on nuImages with 11 classes, FSKD achieves competitive performance across all 16 nuScenes classes after distillation, demonstrating its ability to enable the 3D network to extract generalizable features.

\begin{table}[tb]
        \centering
        \caption{Few-Shot 3D Semantic Segmentation Results}
        \setlength{\tabcolsep}{16pt}
        \label{tab:few-shot}
        \begin{tabular}{@{}lcccc@{}}
                \toprule
                \multirow{2}{*}{Method}               & \multicolumn{2}{c}{nuScenes} & \multicolumn{2}{c}{KITTI}                                       \\
                \cmidrule(lr){2-3} \cmidrule(lr){4-5} &
                1\%                                   & 100\%                        & 1\%                       & 100\%                               \\

                \midrule

                Random                                & 31.8                         & 74.2                      & 39.5             & 53.1             \\
                I2P-MAE~\cite{I2P-MAE}                & 34.2                         & 74.2                      & 40.9             & 53.0             \\
                PPKT~\cite{48}                        & 37.8                         & 74.4                      & 44.2             & 52.9             \\
                SLidR~\cite{51}                       & 38.3                         & 74.7                      & 44.6             & 53.2             \\
                UDAKD (Ours)                          & \underline{39.3}             & \textbf{75.0}             & \underline{45.1} & \textbf{54.0}    \\
                FSKD (Ours)                           & \textbf{44.7}                & \underline{74.9}          & \textbf{46.2}    & \underline{53.8} \\
                \bottomrule
        \end{tabular}
\end{table}

\begin{figure}[tb]
        \centering
        \includegraphics[scale=1]{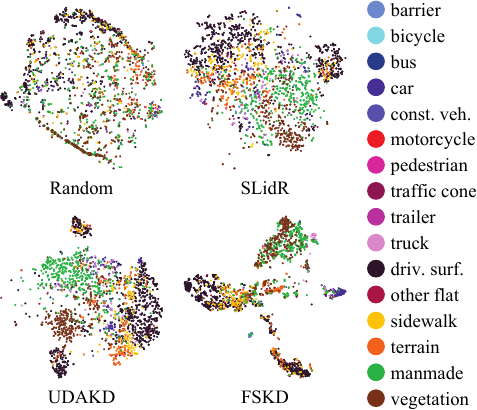}
        \caption{t-SNE visualization of 3D feature extractor outputs pretrained with various methods.}
        \label{fig:t-sne}
\end{figure}

\paragraph{Visual inspection}
We input 3D data from the nuScenes validation set into feature extractors $h_{\omega_h}$ pretrained with various methods to generate feature vectors. These vectors undergo t-SNE dimensionality reduction, with the resulting points color-coded by ground truth semantic labels, as shown in Fig.~\ref{fig:t-sne}. Compared to SLidR~\cite{51}, UDAKD shows clearer separation among dominant categories, such as the deep purple drivable surface, deep green manmade, orange terrain, and brown vegetation, with tighter intra-category clustering. FSKD, which utilizes 2D image labels during 2D network training, shows further improved 3D feature extraction performance.

\subsection{Zero-Shot Semantic Segmentation}

In the zero-shot segmentation setup of FSKD, the 2D classifier $d_{\theta_d}$ trained on images is reused as the 3D classifier $k_{\omega_k}$ with identical structure and parameters. Although not strictly necessary for zero-shot segmentation, this sharing minimally affects training effectiveness and is crucial for zero-shot domain adaptation, discussed in the next subsection. Other processes and hyperparameters remain consistent with those used in few-shot semantic segmentation.

\paragraph{Baselines}
Training 3D networks without direct 3D supervision is largely unexplored. To establish baselines, we adapt OpenScene~\cite{OpenScene}, 2DPASS~\cite{50}, and MVF~\cite{97}. OpenScene uses cosine similarity loss to align 3D features with 2D-3D fusion features. To ensure comparability with other baselines, we train the 2D network with supervised data from the singlemodal dataset $\mathbf{S}$ in the source domain, instead of text-image pairs used in the original paper~\cite{OpenScene}. In MVF, the 2D network generates pseudo-labels, assigning each 3D point the corresponding 2D semantic label if aligned with a 2D pixel. For 2DPASS, lacking ground truth 3D labels, 2D knowledge is transferred to 3D in two steps: fusion features are generated using the original 2DPASS approach but limited to a single scale, and combined with 2D and 3D features as inputs to a fixed classifier trained on 2D images, which outputs semantic logit maps. The total loss includes KL divergence between the fused soft labels and both the 2D and 3D soft labels.

\paragraph{Evaluation results}
We evaluate mIoU for the 11 common semantic classes in nuImages and nuScenes on the nuScenes validation set without fine-tuning the 3D network. As shown in Table~\ref{tab:zero-shot}, these results are not directly comparable to the few-shot segmentation results in Table~\ref{tab:few-shot}. The cosine similarity used for feature alignment, as seen in OpenScene, proves less effective in our context. Our method outperforms the pseudo-label approach of MVF by 7.9 percentage points. While 2DPASS shows a slight improvement over MVF ($+1.0$ mIoU), it still trails our FSKD method by 6.9 mIoU points.

\begin{table}[tb]
        \centering
        \caption{Zero-Shot 3D Semantic Segmentation Results}
        \label{tab:zero-shot}
        \setlength{\tabcolsep}{86pt}
        \begin{tabular}{@{}ll@{}}
                \toprule
                Method                     & mIoU          \\
                \midrule
                OpenScene~\cite{OpenScene} & 41.5          \\
                MVF~\cite{97}              & 55.8          \\
                2DPASS~\cite{50}           & 56.8          \\
                FSKD (Ours)                & \textbf{63.7} \\
                \bottomrule
        \end{tabular}
\end{table}

\subsection{Further Insights into FSKD}

\paragraph{Annotation efficiency}
A key question is whether the performance of FSKD results from overfitting the 3D student to the 2D teacher's outputs. To explore this, we fine-tune SLidR~\cite{51} and FSKD-pretrained 3D networks using different proportions of 3D labels from the nuScenes training set. We evaluate the mIoU on the validation set for the 11 semantic classes, and the results are summarized in Table~\ref{tab:annotation efficiency}. As annotation proportions increase, both methods improve; however, even with $100\%$ annotation availability, the FSKD-pretrained network outperforms SLidR by 0.4 percentage points. This indicates that FSKD not only performs well in zero-shot learning but also enhances the 3D network's capacity to extract generalizable features.

\begin{table}[tb]
        \caption{Segmentation Results with Varying Annotation Availability}
        \label{tab:annotation efficiency}
        \centering
        \setlength{\tabcolsep}{9.2pt}
        \begin{tabular}{@{}lcccccc@{}}
                \toprule
                Method          & 0\%           & 1\%           & 5\%           & 10\%          & 25\%          & 100\%         \\
                \midrule
                SLidR~\cite{51} & 4.1           & 32.9          & 58.2          & 66.6          & \textbf{71.2} & 78.1          \\
                FSKD (Ours)     & \textbf{63.7} & \textbf{62.0} & \textbf{66.3} & \textbf{69.6} & \textbf{71.2} & \textbf{78.6} \\
                \bottomrule
        \end{tabular}
\end{table}

\paragraph{Zero-shot domain adaptation}\label{sec:Zero-Shot Domain Adaptation}
Our approach enables zero-shot domain adaptation by sharing the same classifier between 3D and 2D networks. After knowledge distillation, the 2D network $f_{\theta}$ is retrained on nuImages without mapping its 25 semantic classes and 1 noise class to 11 semantic classes. The noise and ego-vehicle classes are treated as noise, using the remaining 24 classes without merging. During training, we freeze the 2D feature extractor $e_{\theta_e}$ and replace its classifier with a new one, $d_{\theta_d^\prime}^\prime$, forming SegNet*. We then link $d_{\theta_d^\prime}^\prime$ to the previously distilled 3D feature extractor $h_{\omega_h}$ and domain adaptation module $m_{\lambda}$, forming FSKD*, denoted as $g_{\omega^\prime}^\prime = d_{\theta_d^\prime}^\prime \circ m_\lambda \circ h_{\omega_h}$. With closely aligned 3D and 2D features during FSKD, $g_{\omega^\prime}^\prime$ can directly recognize new classes. Table~\ref{tab:zero-shot-da} shows the mIoU for FSKD* and SegNet* across various classes. The 2D network distinguishes debris, pushable/pullable objects, and bicycle racks (previously classified as noise), as well as adult pedestrians and construction workers (previously grouped as pedestrians). The 3D network, sharing the same classifier, differentiates these classes without any parameter adjustments or additional 3D annotations.

\begin{table}[tb]
        \caption{Zero-Shot Domain Adaptation Results}
        \label{tab:zero-shot-da}
        \centering
        \setlength{\tabcolsep}{5.7pt}
        \begin{tabular}{@{}lccccc@{}}
                \toprule

                \multirow{2}{*}{Method}               & \multicolumn{3}{c}{noise} & \multicolumn{2}{c}{pedestrian}                              \\
                \cmidrule(lr){2-4} \cmidrule(lr){5-6} &
                debris                                & pshb. pllb. obj.          & bicycle rack                   & adult & const. wkr.        \\

                \midrule

                FSKD*                                 & 0.5                       & 2.0                            & 6.7   & 47.9        & 10.6 \\
                SegNet*                               & 2.5                       & 9.1                            & 6.3   & 76.0        & 11.7 \\

                \bottomrule
        \end{tabular}
\end{table}

\subsection{Ablation Study}

\paragraph{UDAKD}
In this study, we evaluate UDAKD by omitting the domain adaptation module $m_{\lambda}$ and superpixels. The results in Table~\ref{tab:ablation-da} demonstrate that incorporating $m_{\lambda}$ not only boosts performance but also reduces dependency on superpixels. Even without superpixels, our UDAKD still outperforms SLidR~\cite{51}, which considers the introduction of superpixels as its major contribution.

\begin{table}[tb]
        \centering
        \caption{Ablation Study of UDAKD}
        \label{tab:ablation-da}
        \setlength{\tabcolsep}{21.5pt}
        \begin{tabular}{@{}l c c l@{}}
                \toprule
                Method                                & Sp.            & DA             & mIoU        \\
                \midrule
                UDAKD                                 & \CheckmarkBold & \CheckmarkBold & 39.3        \\
                UDAKD \footnotesize w/o sp.           & \XSolidBrush   & \CheckmarkBold & 38.6 (-0.7) \\
                SLidR~\cite{51}                       & \CheckmarkBold & \XSolidBrush   & 38.3 (-1.0) \\
                SLidR~\cite{51} \footnotesize w/o sp. & \XSolidBrush   & \XSolidBrush   & 36.6 (-2.7) \\
                \bottomrule
        \end{tabular}
\end{table}

\paragraph{FSKD}
The FSKD ablation study measures the mIoU of the 3D network trained on the nuScenes training set via crossmodal knowledge distillation, without additional fine-tuning using 3D ground truth labels. Table~\ref{tab:ablation-fskd} highlights the importance of each component in UDAKD. Removing semantic-based distillation reduces performance by 15.7 percentage points, underscoring the importance of aligning semantic soft labels for zero-shot crossmodal supervision. The domain adaptation module is essential for isolating modality-specific information and aligning 2D and 3D outputs at the feature level, enabling effective knowledge distillation at both feature and semantic levels.

\begin{table}[tb]
        \centering
        \caption{Ablation Study of FSKD}
        \label{tab:ablation-fskd}
        \centering
        \setlength{\tabcolsep}{52pt}
        \begin{tabular}{@{}ll@{}}
                \toprule
                Method                                   & mIoU         \\
                \midrule
                FSKD                                     & 63.7         \\
                FSKD \footnotesize w/o soft labels       & 55.1 (-8.6)  \\
                FSKD \footnotesize w/o feature-based KD  & 55.3 (-8.4)  \\
                FSKD \footnotesize w/o semantic-based KD & 48.0 (-15.7) \\
                FSKD \footnotesize w/o DA module         & 48.5 (-15.2) \\
                \bottomrule
        \end{tabular}
\end{table}

\begin{figure}[tb]
        \centering
        \includegraphics[scale=1]{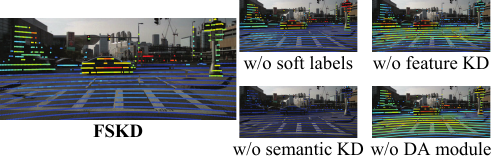}
        \caption{Visualization of 3D features with color-coded $\ell^2$-norm.}
        \label{fig:ablation-visualization}
\end{figure}

Fig.~\ref{fig:ablation-visualization} shows the feature map outputs of the 3D feature extractor $h_{\omega_h}$ with different components ablated. FSKD effectively distinguishes between classes and maintains consistency within the same class, clearly identifying roads, cars, and buildings. When components are removed, the features' ability to convey semantic information diminishes. Even in the \emph{w/o soft labels} scenario, there is significant risk of confusion, such as mistaking road surfaces color-coded in light blue for cars or erroneously associating cars color-coded in red and yellow with traffic lights.

\section{Conclusion}

This work introduces two crossmodal knowledge distillation methods: \textbf{U}n\-su\-per\-vised \textbf{D}o\-main \textbf{A}d\-ap\-ta\-tion \textbf{K}nowl\-edge \textbf{D}is\-til\-la\-tion (UDAKD) and \textbf{F}ea\-ture and \textbf{S}e\-man\-tic-based \textbf{K}nowl\-edge \textbf{D}is\-til\-la\-tion (FSKD). UDAKD utilizes unlabeled image data for 2D teacher training, while FSKD leverages labeled images. Experimental results demonstrate the superiority of both UDAKD and FSKD over existing state-of-the-art methods.

While we highlight the role of a domain adaptation module in the 3D network, further research is needed to optimize its structure for crossmodal information transfer. Additionally, although 2D-3D correspondence can act as a form of supervision, we classify UDAKD as unsupervised due to the reliable, one-time extrinsic parameter calibration used to establish this correspondence. Accurately determining coordinate transformations between sparse point clouds and images without prior knowledge remains a substantive enigma.

\section*{ACKNOWLEDGMENT}

The work is supported in part by the National Natural Science Foundation of China (No. 62176004, No. U1713217), Intelligent Robotics and Autonomous Vehicle Lab (RAV), the Fundamental Research Funds for the Central Universities, and High-performance Computing Platform of Peking University.

\bibliographystyle{IEEEtran}
\bibliography{ref}

\begin{thebibliography}{10}
\providecommand{\url}[1]{#1}
\csname url@samestyle\endcsname
\providecommand{\newblock}{\relax}
\providecommand{\bibinfo}[2]{#2}
\providecommand{\BIBentrySTDinterwordspacing}{\spaceskip=0pt\relax}
\providecommand{\BIBentryALTinterwordstretchfactor}{4}
\providecommand{\BIBentryALTinterwordspacing}{\spaceskip=\fontdimen2\font plus
\BIBentryALTinterwordstretchfactor\fontdimen3\font minus \fontdimen4\font\relax}
\providecommand{\BIBforeignlanguage}[2]{{%
\expandafter\ifx\csname l@#1\endcsname\relax
\typeout{** WARNING: IEEEtran.bst: No hyphenation pattern has been}%
\typeout{** loaded for the language `#1'. Using the pattern for}%
\typeout{** the default language instead.}%
\else
\language=\csname l@#1\endcsname
\fi
#2}}
\providecommand{\BIBdecl}{\relax}
\BIBdecl

\bibitem{9}
H.~Caesar, V.~Bankiti, A.~H. Lang, S.~Vora, V.~E. Liong, Q.~Xu, A.~Krishnan, Y.~Pan, G.~Baldan, and O.~Beijbom, ``nuscenes: A multimodal dataset for autonomous driving,'' in \emph{Proceedings of the IEEE/CVF conference on computer vision and pattern recognition}, 2020, pp. 11\,621--11\,631.

\bibitem{10}
A.~Geiger, P.~Lenz, C.~Stiller, and R.~Urtasun, ``Vision meets robotics: The kitti dataset,'' \emph{The International Journal of Robotics Research}, vol.~32, no.~11, pp. 1231--1237, 2013.

\bibitem{11}
J.~Houston, G.~Zuidhof, L.~Bergamini, Y.~Ye, L.~Chen, A.~Jain, S.~Omari, V.~Iglovikov, and P.~Ondruska, ``One thousand and one hours: Self-driving motion prediction dataset,'' in \emph{Conference on Robot Learning}.\hskip 1em plus 0.5em minus 0.4em\relax PMLR, 2021, pp. 409--418.

\bibitem{12}
P.~Sun, H.~Kretzschmar, X.~Dotiwalla, A.~Chouard, V.~Patnaik, P.~Tsui, J.~Guo, Y.~Zhou, Y.~Chai, B.~Caine \emph{et~al.}, ``Scalability in perception for autonomous driving: Waymo open dataset,'' in \emph{Proceedings of the IEEE/CVF conference on computer vision and pattern recognition}, 2020, pp. 2446--2454.

\bibitem{41}
K.~Genova, X.~Yin, A.~Kundu, C.~Pantofaru, F.~Cole, A.~Sud, B.~Brewington, B.~Shucker, and T.~Funkhouser, ``Learning 3d semantic segmentation with only 2d image supervision,'' in \emph{2021 International Conference on 3D Vision (3DV)}.\hskip 1em plus 0.5em minus 0.4em\relax IEEE, 2021, pp. 361--372.

\bibitem{51}
C.~Sautier, G.~Puy, S.~Gidaris, A.~Boulch, A.~Bursuc, and R.~Marlet, ``Image-to-lidar self-supervised distillation for autonomous driving data,'' in \emph{Proceedings of the IEEE/CVF Conference on Computer Vision and Pattern Recognition}, 2022, pp. 9891--9901.

\bibitem{102}
L.~Yi, B.~Gong, and T.~Funkhouser, ``Complete \& label: A domain adaptation approach to semantic segmentation of lidar point clouds,'' in \emph{Proceedings of the IEEE/CVF conference on computer vision and pattern recognition}, 2021, pp. 15\,363--15\,373.

\bibitem{48}
Y.-C. Liu, Y.-K. Huang, H.-Y. Chiang, H.-T. Su, Z.-Y. Liu, C.-T. Chen, C.-Y. Tseng, and W.~H. Hsu, ``Learning from 2d: Contrastive pixel-to-point knowledge transfer for 3d pretraining,'' \emph{arXiv preprint arXiv:2104.04687}, 2021.

\bibitem{54}
Z.~Xue, Z.~Gao, S.~Ren, and H.~Zhao, ``The modality focusing hypothesis: Towards understanding crossmodal knowledge distillation,'' \emph{arXiv preprint arXiv:2206.06487}, 2022.

\bibitem{62}
J.-J. Liu, Q.~Hou, M.-M. Cheng, C.~Wang, and J.~Feng, ``Improving convolutional networks with self-calibrated convolutions,'' in \emph{Proceedings of the IEEE/CVF conference on computer vision and pattern recognition}, 2020, pp. 10\,096--10\,105.

\bibitem{53}
Y.~Hong, H.~Dai, and Y.~Ding, ``Cross-modality knowledge distillation network for monocular 3d object detection,'' in \emph{Computer Vision--ECCV 2022: 17th European Conference, Tel Aviv, Israel, October 23--27, 2022, Proceedings, Part X}.\hskip 1em plus 0.5em minus 0.4em\relax Springer, 2022, pp. 87--104.

\bibitem{17}
O.~Ronneberger, P.~Fischer, and T.~Brox, ``U-net: Convolutional networks for biomedical image segmentation,'' in \emph{Medical Image Computing and Computer-Assisted Intervention--MICCAI 2015: 18th International Conference, Munich, Germany, October 5-9, 2015, Proceedings, Part III 18}.\hskip 1em plus 0.5em minus 0.4em\relax Springer, 2015, pp. 234--241.

\bibitem{18}
C.~R. Qi, H.~Su, K.~Mo, and L.~J. Guibas, ``Pointnet: Deep learning on point sets for 3d classification and segmentation,'' in \emph{Proceedings of the IEEE conference on computer vision and pattern recognition}, 2017, pp. 652--660.

\bibitem{19}
C.~R. Qi, L.~Yi, H.~Su, and L.~J. Guibas, ``Pointnet++: Deep hierarchical feature learning on point sets in a metric space,'' \emph{Advances in neural information processing systems}, vol.~30, 2017.

\bibitem{21}
S.~Wang, S.~Suo, W.-C. Ma, A.~Pokrovsky, and R.~Urtasun, ``Deep parametric continuous convolutional neural networks,'' in \emph{Proceedings of the IEEE conference on computer vision and pattern recognition}, 2018, pp. 2589--2597.

\bibitem{22}
H.~Thomas, C.~R. Qi, J.-E. Deschaud, B.~Marcotegui, F.~Goulette, and L.~J. Guibas, ``Kpconv: Flexible and deformable convolution for point clouds,'' in \emph{Proceedings of the IEEE/CVF international conference on computer vision}, 2019, pp. 6411--6420.

\bibitem{27}
B.~Wu, A.~Wan, X.~Yue, and K.~Keutzer, ``Squeezeseg: Convolutional neural nets with recurrent crf for real-time road-object segmentation from 3d lidar point cloud,'' in \emph{2018 IEEE international conference on robotics and automation (ICRA)}.\hskip 1em plus 0.5em minus 0.4em\relax IEEE, 2018, pp. 1887--1893.

\bibitem{28}
B.~Wu, X.~Zhou, S.~Zhao, X.~Yue, and K.~Keutzer, ``Squeezesegv2: Improved model structure and unsupervised domain adaptation for road-object segmentation from a lidar point cloud,'' in \emph{2019 International Conference on Robotics and Automation (ICRA)}.\hskip 1em plus 0.5em minus 0.4em\relax IEEE, 2019, pp. 4376--4382.

\bibitem{29}
C.~Xu, B.~Wu, Z.~Wang, W.~Zhan, P.~Vajda, K.~Keutzer, and M.~Tomizuka, ``Squeezesegv3: Spatially-adaptive convolution for efficient point-cloud segmentation,'' in \emph{Computer Vision--ECCV 2020: 16th European Conference, Glasgow, UK, August 23--28, 2020, Proceedings, Part XXVIII 16}.\hskip 1em plus 0.5em minus 0.4em\relax Springer, 2020, pp. 1--19.

\bibitem{30}
A.~Milioto, I.~Vizzo, J.~Behley, and C.~Stachniss, ``Rangenet++: Fast and accurate lidar semantic segmentation,'' in \emph{2019 IEEE/RSJ international conference on intelligent robots and systems (IROS)}.\hskip 1em plus 0.5em minus 0.4em\relax IEEE, 2019, pp. 4213--4220.

\bibitem{31}
T.~Cortinhal, G.~Tzelepis, and E.~Erdal~Aksoy, ``Salsanext: Fast, uncertainty-aware semantic segmentation of lidar point clouds,'' in \emph{Advances in Visual Computing: 15th International Symposium, ISVC 2020, San Diego, CA, USA, October 5--7, 2020, Proceedings, Part II 15}.\hskip 1em plus 0.5em minus 0.4em\relax Springer, 2020, pp. 207--222.

\bibitem{32}
Y.~Zhang, Z.~Zhou, P.~David, X.~Yue, Z.~Xi, B.~Gong, and H.~Foroosh, ``Polarnet: An improved grid representation for online lidar point clouds semantic segmentation,'' in \emph{Proceedings of the IEEE/CVF Conference on Computer Vision and Pattern Recognition}, 2020, pp. 9601--9610.

\bibitem{33}
Z.~Zhou, Y.~Zhang, and H.~Foroosh, ``Panoptic-polarnet: Proposal-free lidar point cloud panoptic segmentation,'' in \emph{Proceedings of the IEEE/CVF Conference on Computer Vision and Pattern Recognition}, 2021, pp. 13\,194--13\,203.

\bibitem{35}
C.~Choy, J.~Gwak, and S.~Savarese, ``4d spatio-temporal convnets: Minkowski convolutional neural networks,'' in \emph{Proceedings of the IEEE/CVF conference on computer vision and pattern recognition}, 2019, pp. 3075--3084.

\bibitem{36}
S.~Xie, J.~Gu, D.~Guo, C.~R. Qi, L.~Guibas, and O.~Litany, ``Pointcontrast: Unsupervised pre-training for 3d point cloud understanding,'' in \emph{Computer Vision--ECCV 2020: 16th European Conference, Glasgow, UK, August 23--28, 2020, Proceedings, Part III 16}.\hskip 1em plus 0.5em minus 0.4em\relax Springer, 2020, pp. 574--591.

\bibitem{37}
X.~Zhu, H.~Zhou, T.~Wang, F.~Hong, Y.~Ma, W.~Li, H.~Li, and D.~Lin, ``Cylindrical and asymmetrical 3d convolution networks for lidar segmentation,'' in \emph{Proceedings of the IEEE/CVF conference on computer vision and pattern recognition}, 2021, pp. 9939--9948.

\bibitem{38}
R.~Cheng, R.~Razani, E.~Taghavi, E.~Li, and B.~Liu, ``(af)2-s3net: Attentive feature fusion with adaptive feature selection for sparse semantic segmentation network,'' in \emph{Proceedings of the IEEE/CVF conference on computer vision and pattern recognition}, 2021, pp. 12\,547--12\,556.

\bibitem{39}
H.~Tang, Z.~Liu, S.~Zhao, Y.~Lin, J.~Lin, H.~Wang, and S.~Han, ``Searching efficient 3d architectures with sparse point-voxel convolution,'' in \emph{Computer Vision--ECCV 2020: 16th European Conference, Glasgow, UK, August 23--28, 2020, Proceedings, Part XXVIII}.\hskip 1em plus 0.5em minus 0.4em\relax Springer, 2020, pp. 685--702.

\bibitem{100}
X.~Chen, S.~Xu, X.~Zou, T.~Cao, D.-Y. Yeung, and L.~Fang, ``Svqnet: Sparse voxel-adjacent query network for 4d spatio-temporal lidar semantic segmentation,'' in \emph{Proceedings of the IEEE/CVF International Conference on Computer Vision}, 2023, pp. 8569--8578.

\bibitem{45}
S.~Gupta, J.~Hoffman, and J.~Malik, ``Cross modal distillation for supervision transfer,'' in \emph{Proceedings of the IEEE conference on computer vision and pattern recognition}, 2016, pp. 2827--2836.

\bibitem{47}
J.~Meyer, A.~Eitel, T.~Brox, and W.~Burgard, ``Improving unimodal object recognition with multimodal contrastive learning,'' in \emph{2020 IEEE/RSJ International Conference on Intelligent Robots and Systems (IROS)}.\hskip 1em plus 0.5em minus 0.4em\relax IEEE, 2020, pp. 5656--5663.

\bibitem{I2P-MAE}
R.~Zhang, L.~Wang, Y.~Qiao, P.~Gao, and H.~Li, ``Learning 3d representations from 2d pre-trained models via image-to-point masked autoencoders,'' in \emph{Proceedings of the IEEE/CVF Conference on Computer Vision and Pattern Recognition}, 2023, pp. 21\,769--21\,780.

\bibitem{99}
P.~Cong, Y.~Xu, Y.~Ren, J.~Zhang, L.~Xu, J.~Wang, J.~Yu, and Y.~Ma, ``Weakly supervised 3d multi-person pose estimation for large-scale scenes based on monocular camera and single lidar,'' in \emph{Proceedings of the AAAI Conference on Artificial Intelligence}, vol.~37, 2023, pp. 461--469.

\bibitem{42}
M.~Jaritz, T.-H. Vu, R.~d. Charette, E.~Wirbel, and P.~P{\'e}rez, ``xmuda: Cross-modal unsupervised domain adaptation for 3d semantic segmentation,'' in \emph{Proceedings of the IEEE/CVF conference on computer vision and pattern recognition}, 2020, pp. 12\,605--12\,614.

\bibitem{49}
M.~Jaritz, T.-H. Vu, R.~De~Charette, {\'E}.~Wirbel, and P.~P{\'e}rez, ``Cross-modal learning for domain adaptation in 3d semantic segmentation,'' \emph{IEEE Transactions on Pattern Analysis and Machine Intelligence}, vol.~45, no.~2, pp. 1533--1544, 2022.

\bibitem{OpenScene}
S.~Peng, K.~Genova, C.~Jiang, A.~Tagliasacchi, M.~Pollefeys, T.~Funkhouser \emph{et~al.}, ``Openscene: 3d scene understanding with open vocabularies,'' in \emph{Proceedings of the IEEE/CVF Conference on Computer Vision and Pattern Recognition}, 2023, pp. 815--824.

\bibitem{52}
R.~Achanta, A.~Shaji, K.~Smith, A.~Lucchi, P.~Fua, and S.~S{\"u}sstrunk, ``Slic superpixels compared to state-of-the-art superpixel methods,'' \emph{IEEE transactions on pattern analysis and machine intelligence}, vol.~34, no.~11, pp. 2274--2282, 2012.

\bibitem{55}
C.~Xu, S.~Yang, T.~Galanti, B.~Wu, X.~Yue, B.~Zhai, W.~Zhan, P.~Vajda, K.~Keutzer, and M.~Tomizuka, ``Image2point: 3d point-cloud understanding with 2d image pretrained models,'' in \emph{Computer Vision--ECCV 2022: 17th European Conference, Tel Aviv, Israel, October 23--27, 2022, Proceedings, Part XXXVII}.\hskip 1em plus 0.5em minus 0.4em\relax Springer, 2022, pp. 638--656.

\bibitem{56}
T.~Galanti, A.~Gy{\"o}rgy, and M.~Hutter, ``On the role of neural collapse in transfer learning,'' \emph{arXiv preprint arXiv:2112.15121}, 2021.

\bibitem{86}
Y.~Huang, W.~Zheng, Y.~Zhang, J.~Zhou, and J.~Lu, ``Tri-perspective view for vision-based 3d semantic occupancy prediction,'' in \emph{Proceedings of the IEEE/CVF Conference on Computer Vision and Pattern Recognition}, 2023, pp. 9223--9232.

\bibitem{68}
V.~Badrinarayanan, A.~Kendall, and R.~Cipolla, ``Segnet: A deep convolutional encoder-decoder architecture for image segmentation,'' \emph{IEEE transactions on pattern analysis and machine intelligence}, vol.~39, no.~12, pp. 2481--2495, 2017.

\bibitem{43}
J.~Li, H.~Dai, H.~Han, and Y.~Ding, ``Mseg3d: Multi-modal 3d semantic segmentation for autonomous driving,'' in \emph{Proceedings of the IEEE/CVF Conference on Computer Vision and Pattern Recognition}, 2023, pp. 21\,694--21\,704.

\bibitem{88}
K.~El~Madawi, H.~Rashed, A.~El~Sallab, O.~Nasr, H.~Kamel, and S.~Yogamani, ``Rgb and lidar fusion based 3d semantic segmentation for autonomous driving,'' in \emph{2019 IEEE Intelligent Transportation Systems Conference (ITSC)}.\hskip 1em plus 0.5em minus 0.4em\relax IEEE, 2019, pp. 7--12.

\bibitem{89}
G.~P. Meyer, J.~Charland, D.~Hegde, A.~Laddha, and C.~Vallespi-Gonzalez, ``Sensor fusion for joint 3d object detection and semantic segmentation,'' in \emph{Proceedings of the IEEE/CVF conference on computer vision and pattern recognition workshops}, 2019, pp. 0--0.

\bibitem{90}
S.~Vora, A.~H. Lang, B.~Helou, and O.~Beijbom, ``Pointpainting: Sequential fusion for 3d object detection,'' in \emph{Proceedings of the IEEE/CVF conference on computer vision and pattern recognition}, 2020, pp. 4604--4612.

\bibitem{91}
G.~Krispel, M.~Opitz, G.~Waltner, H.~Possegger, and H.~Bischof, ``Fuseseg: Lidar point cloud segmentation fusing multi-modal data,'' in \emph{Proceedings of the IEEE/CVF winter conference on applications of computer vision}, 2020, pp. 1874--1883.

\bibitem{92}
Z.~Xue, X.~Yu, X.~Tan, B.~Liu, A.~Yu, and X.~Wei, ``Multiscale deep learning network with self-calibrated convolution for hyperspectral and lidar data collaborative classification,'' \emph{IEEE Transactions on Geoscience and Remote Sensing}, vol.~60, pp. 1--16, 2021.

\bibitem{87}
A.~v.~d. Oord, Y.~Li, and O.~Vinyals, ``Representation learning with contrastive predictive coding,'' \emph{arXiv preprint arXiv:1807.03748}, 2018.

\bibitem{44}
J.~Behley, M.~Garbade, A.~Milioto, J.~Quenzel, S.~Behnke, C.~Stachniss, and J.~Gall, ``Semantickitti: A dataset for semantic scene understanding of lidar sequences,'' in \emph{Proceedings of the IEEE/CVF international conference on computer vision}, 2019, pp. 9297--9307.

\bibitem{58}
J.~Deng, W.~Dong, R.~Socher, L.-J. Li, K.~Li, and L.~Fei-Fei, ``Imagenet: A large-scale hierarchical image database,'' in \emph{2009 IEEE conference on computer vision and pattern recognition}.\hskip 1em plus 0.5em minus 0.4em\relax Ieee, 2009, pp. 248--255.

\bibitem{64}
X.~Chen, H.~Fan, R.~Girshick, and K.~He, ``Improved baselines with momentum contrastive learning,'' \emph{arXiv preprint arXiv:2003.04297}, 2020.

\bibitem{103}
A.~Krizhevsky, G.~Hinton \emph{et~al.}, ``Learning multiple layers of features from tiny images,'' \emph{Master's thesis, University of Toronto}, 2009.

\bibitem{50}
X.~Yan, J.~Gao, C.~Zheng, C.~Zheng, R.~Zhang, S.~Cui, and Z.~Li, ``2dpass: 2d priors assisted semantic segmentation on lidar point clouds,'' in \emph{Computer Vision--ECCV 2022: 17th European Conference, Tel Aviv, Israel, October 23--27, 2022, Proceedings, Part XXVIII}.\hskip 1em plus 0.5em minus 0.4em\relax Springer, 2022, pp. 677--695.

\bibitem{97}
A.~Hermans, G.~Floros, and B.~Leibe, ``Dense 3d semantic mapping of indoor scenes from rgb-d images,'' in \emph{2014 IEEE International Conference on Robotics and Automation (ICRA)}.\hskip 1em plus 0.5em minus 0.4em\relax IEEE, 2014, pp. 2631--2638.

\end{thebibliography}

\end{document}